\documentclass[10pt,twocolumn,letterpaper]{article}
\PassOptionsToPackage{numbers,compress}{natbib}
\usepackage{iccv}              

\definecolor{iccvblue}{rgb}{0.21,0.49,0.74}
\usepackage[pagebackref,breaklinks,colorlinks,allcolors=iccvblue]{hyperref}

\usepackage[numbers]{natbib}

\usepackage{times}

\usepackage{graphicx}
\usepackage{amsmath}
\usepackage{amssymb}
\usepackage{booktabs}
\usepackage{cuted}
\usepackage{duckuments}
\usepackage{algorithm}
\usepackage{algcompatible}
\usepackage{amsmath}
\usepackage{tocloft}

\usepackage[capitalize]{cleveref}
\Crefname{section}{Sec.}{Secs.}
\crefname{section}{Sec.}{Secs.}
\Crefname{table}{Tab.}{Tabs.}
\crefname{table}{Tab.}{Tabs.}
\Crefname{figure}{Fig.}{Figs.}
\crefname{figure}{Fig.}{Figs.}
\Crefname{appendix}{App.}{Apps.}
\crefname{appendix}{App.}{Apps.}

\usepackage[utf8]{inputenc} %
\usepackage[T1]{fontenc}    %
\usepackage{url}            %
\usepackage{amsfonts}       %
\usepackage{nicefrac}       %
\usepackage[nopatch=eqnum]{microtype}      %

\usepackage{epsfig}
\usepackage{tabularx}
\usepackage{bbm}
\usepackage{color}
\usepackage{wrapfig}
\usepackage{subcaption}
\usepackage{float}
\usepackage{makecell}
\usepackage[percent]{overpic}
\usepackage{adjustbox}

\usepackage{multirow}
\usepackage{inconsolata}
\usepackage{comment}
\usepackage{xspace}

\newcommand{\stdhide}[1]{}

\newcolumntype{Y}{>{\centering\arraybackslash}X}

\newcommand{\ourabbr}{PhysTwin\xspace}

\title{PhysTwin: Physics-Informed Reconstruction and Simulation of \\ Deformable Objects from Videos}

\author{Hanxiao Jiang$^{1, 2}$\quad Hao-Yu Hsu$^{2}$\quad Kaifeng Zhang$^1$\quad Hsin-Ni Yu$^2$\quad Shenlong Wang$^2$\quad Yunzhu Li$^1$ \\\small{$^1$Columbia University\quad $^2$University of Illinois Urbana-Champaign}}

\addtocontents{toc}{\setcounter{tocdepth}{-10}}

\begin{document}
\maketitle
\begin{strip}
    \centering
    \vspace{-55pt}
    \includegraphics[width=\linewidth]{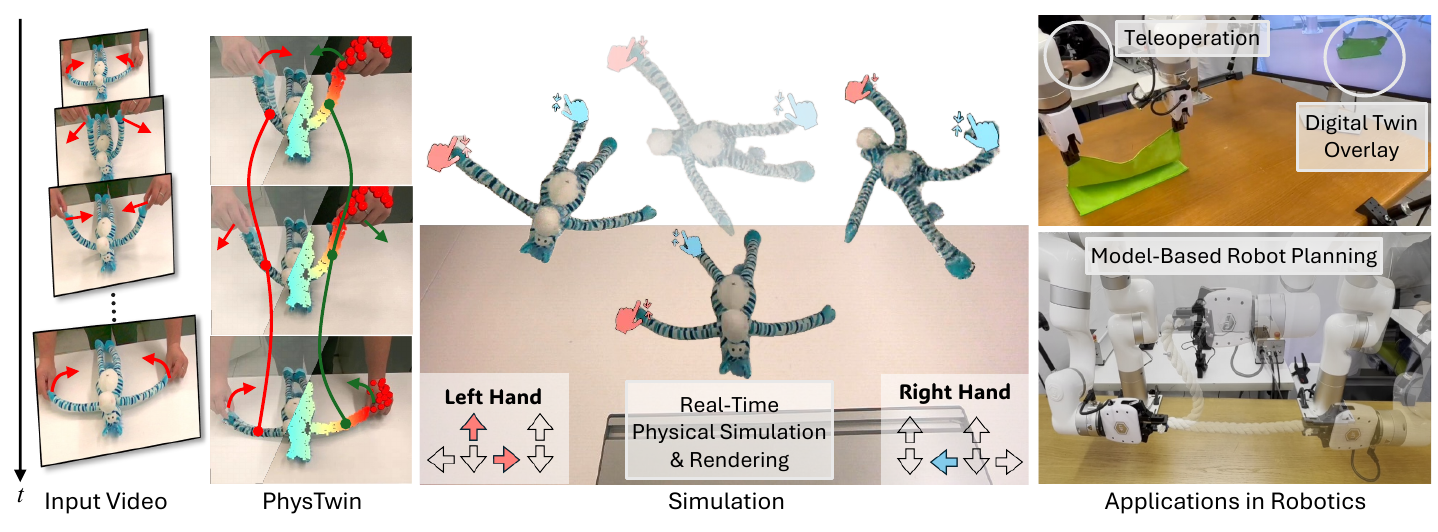
    }
    \vspace{-25pt}
    \captionof{figure}{\small
    \textbf{\ourabbr} takes sparse videos (three camera views) of deformable objects under interaction as input and reconstructs a simulatable digital twin with complete geometry, high-fidelity appearance, and accurate physical parameters. This enables multiple applications, such as real-time interactive simulation using keyboards and robotic teleoperation devices, as well as model-based robot planning.}
    \vspace{-10pt}
    \label{fig:teaser}
\end{strip}

\begin{abstract}
Creating a physical digital twin of a real-world object has immense potential in robotics, content creation, and XR. 
In this paper, we present \ourabbr, a novel framework that uses sparse videos of dynamic objects under interaction to produce a photo- and physically realistic, real-time interactive virtual replica.
Our approach centers on two key components: (1) a physics-informed representation that combines spring-mass models for realistic physical simulation, generative shape models for geometry, and Gaussian splats for rendering; and (2) a novel multi-stage, optimization-based inverse modeling framework that reconstructs complete geometry, infers dense physical properties, and replicates realistic appearance from videos. 
Our method integrates an inverse physics framework with visual perception cues, enabling high-fidelity reconstruction even from partial, occluded, and limited viewpoints.
\ourabbr~supports modeling various deformable objects, including ropes, stuffed animals, cloth, and delivery packages. 
Experiments show that \ourabbr~outperforms competing methods in reconstruction, rendering, future prediction, and simulation under novel interactions. We further demonstrate its applications in interactive real-time simulation and model-based robotic motion planning. Project Page: \url{https://jianghanxiao.github.io/phystwin-web/}

\end{abstract}

\vspace{-20pt}
\section{Introduction}

The construction of interactive digital twins is essential for modeling the world and simulating future states, with applications in virtual reality, augmented reality, and robotic manipulation. A physically realistic digital twin (\ourabbr) should accurately capture the geometry, appearance, and physical properties of an object, allowing simulations that closely match observations in the real world. However, constructing such a representation from sparse observations remains a significant challenge.

The creation of digital twins for deformable objects has long been a challenging topic in the vision community. 
While dynamic 3D methods (e.g., dynamic NeRFs~\cite{driess2023learning, li2023pac, li20223d, li2021neural, li2023dynibar, park2021nerfies, park2021hypernerf, pumarola2021d, tretschk2021non, wang2023flow, guo2023forward, cao2023hexplane, fridovich2023k, gao2022monocular, xian2021space, tretschk2021nonrigid, chu2022physics, peng2021CageNeRF}, dynamic 3D Gaussians~\cite{luiten2024dynamic, wu20244d, yang2024deformable, huang2024sc, kratimenos2024dynmf, lin2024gaussian, yu2024cogs, yang2023real, duan20244d}) capture observed motion, appearance, and geometry from videos, they omit the underlying physics and are thus unsuitable for simulating outcomes in unseen interactions. 
While recent neural-based models~\cite{wu2019learning, ma2023learning, xu2019densephysnet, evans2022context, chen2022comphy, shi2024robocraft, shi2023robocook, pfaff2020learning, lin2022learning, li2018learning, sanchez2020learning, zhang2024dynamic} learn intuitive physics models from videos, they require large amounts of data and remain limited to specific objects or motions, whereas physics-driven approaches~\cite{zhang2024physdreamer, zhong2024reconstruction, xie2024physgaussian, feng2024pie, li2023pac, Qiao2021Differentiable, du2021diffpd} often rely on pre-scanned shapes or dense observations to mitigate ill-posedness.
Additionally, it requires dense viewpoint coverage and supports only limited motion types, making it unsuitable for general dynamics modeling.

In this work, we aim to build an interactive \ourabbr from sparse-viewpoint RGB-D video sequences, capturing object geometry, non-rigid dynamic physics, and appearance for realistic physical simulation and rendering. 
We model deformable object dynamics with a spring-mass-based representation, enabling efficient physical simulation and handling a wide range of common objects, such as ropes, stuffed animals, cloth, and delivery packages.
To address the challenges posed by sparse observations, we leverage shape priors and motion estimation from advanced 3D generative models~\cite{xiang2024structured} and vision foundation models~\cite{ren2024grounded, karaev2024cotracker3, rombach2022high} to estimate the topology, geometry, and physical parameters of our physical representation. 
Since some physical parameters (such as topology-related properties) are non-differentiable and optimizing them efficiently is non-trivial, we design a hierarchical sparse-to-dense optimization strategy. This strategy integrates zero-order optimization~\cite{hansen2006cma} for non-differentiable topology and sparse physical parameters (e.g., collision parameters and homogeneous spring stiffness), while employing first-order gradient-based optimization to refine dense spring stiffness and further optimize collision parameters.
For appearance modeling, we adopt a Gaussian blending strategy, initializing static Gaussians from sparse observations in the first frame using shape priors and deforming them with a linear blending algorithm to generate realistic dynamic appearances.

Our inverse modeling framework effectively constructs interactive \ourabbr from videos of objects under interaction. We create a real-world deformable object interaction dataset and evaluate our method on three key tasks: reconstruction and resimulation, future prediction, and generalization to unseen interactions. Both quantitative and qualitative results demonstrate that our reconstructed \ourabbr aligns accurately with real-world observations, achieves precise future predictions, and generates realistic simulations under diverse unseen interactions. Furthermore, the high computational efficiency of our physics simulator enables real-time dynamics and rendering of our constructed \ourabbr, facilitating multiple applications, including real-time interactive simulation and model-based robotic motion planning.

\begin{figure*}
    \centering
    \vspace{-20pt}
    \includegraphics[width=\linewidth]{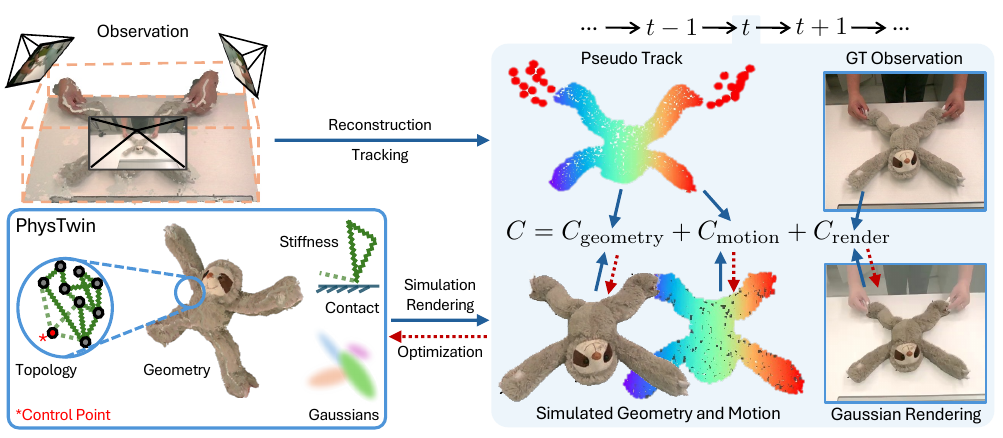}
    \vspace{-20pt}
    \captionof{figure}{\small
    \textbf{Overview of Our \ourabbr Framework.} We present an overview of our PhysTwin framework, where the core representation includes geometry, topology, physical parameters (associated with springs and contacts), and Gaussian kernels. To optimize PhysTwin, we minimize the rendering loss and the discrepancy between simulated and observed geometry/motion. The rendering loss optimizes the Gaussian kernels, while the geometry and motion losses refine the overall geometry, topology, and physical parameters in PhysTwin.
    } 
    \label{fig:pipeline}
    \vspace{-15pt}
\end{figure*}

\section{Related Works}

\textbf{Dynamic Scene Reconstruction.} 
Dynamic scene reconstruction aims to recover the underlying representation of dynamic scenes from inputs like depth scans~\cite{curless1996volumetric, li2008global}, RGBD videos~\cite{newcombe2015dynamicfusion}, or monocular or multi-view videos~\cite{attal2023hyperreel, kratimenos2024dynmf, li2023dynibar, luiten2024dynamic, park2021nerfies, park2021hypernerf, pumarola2021d, wang2023flow, xian2021space, yu2023dylin, yu2024cogs, tretschk2021nonrigid, chu2022physics}. Recent advancements in dynamic scene modeling have involved the adaptation of novel scene representations, including Neural Radiance Fields (NeRF)~\cite{guan2022neurofluid, driess2023learning, li2023pac, li20223d, li2021neural, li2023dynibar, park2021nerfies, park2021hypernerf, pumarola2021d, tretschk2021non, wang2023flow, guo2023forward, cao2023hexplane, fridovich2023k, gao2022monocular, li2021neural, xian2021space, tretschk2021nonrigid, chu2022physics, peng2021CageNeRF} and 3D Gaussian splats~\cite{luiten2024dynamic, wu20244d, yang2024deformable, huang2024sc, kratimenos2024dynmf, lin2024gaussian, yu2024cogs, yang2023real, duan20244d}. D-NeRF~\cite{pumarola2021d} extends a canonical NeRF on dynamic scenes by optimizing a deformable field. Similarly, Deformable 3D-GS~\cite{yang2024deformable} optimizes a deformation field of each Gaussian kernel. Dynamic 3D-GS~\cite{luiten2024dynamic} optimizes the motion of Gaussian kernels for each frame to capture scene dynamics. 4D-GS~\cite{wu20244d} modulates 3D Gaussians with 4D neural voxels for dynamic multi-view synthesis. Although these methods achieve high-fidelity results in dynamic multi-view synthesis, they primarily focus on reconstructing scene appearance and geometry without capturing real-world dynamics, limiting their ability to support action-conditioned future predictions and interactive simulations.

\textbf{Physics-Based Simulation of Deformable Objects.} 
Another line of work incorporates physical simulators to perform system identification of physical parameters during reconstruction. Earlier methods relied on pre-scanned static objects and required clean point cloud observations~\cite{wang2015deformation, Qiao2021Differentiable, du2021diffpd, rojas2021differentiable, geilinger2020add, heiden2021disect, jatavallabhula2021gradsim, ma2022risp}. Most recent approaches build upon SDF~\cite{qiao2022neuphysics}, NeRF~\cite{feng2024pie, li2023pac, chen2022virtual} or Gaussian Splatting~\cite{zhang2024physdreamer, zhong2024reconstruction, xie2024physgaussian, jiang2024vr} to support more flexible physical digital twin reconstruction. 
Several works~\cite{feng2024pie, jiang2024vr, xie2024physgaussian} manually specify physics parameters, resulting in a mismatch between the simulation and real-world video observations. Other works~\cite{zhang2024physdreamer, li2023pac, chen2022virtual, zhong2024reconstruction, qiao2022neuphysics} attempt to estimate physical parameters from videos. However, they are often constrained to synthetic data, limited motion, or the need for dense viewpoints to accurately reconstruct static geometry, limiting their practical applicability.
The closest related work to ours is Spring-Gaus~\cite{zhong2024reconstruction}, which also utilizes a 3D Spring-Mass model for learning from videos. However, their physical model is overly regularized and violates real-world physics, lacking momentum conservation and realistic gravity. Moreover, Spring-Gaus requires dense viewpoint coverage to reconstruct the full geometry at the initial state, which is impractical in many real-world settings. The motions are also limited to tabletop collisions and lack action inputs, making Spring-Gaus unsuitable as a general dynamics model for downstream applications.

\textbf{Learning-Based Simulation of Deformable Objects.} 
Analytically modeling the dynamics of deformable objects is challenging due to the high complexity of the state space and the variability of physical properties. Recent works~\cite{wu2019learning, ma2023learning, xu2019densephysnet, evans2022context, chen2022comphy} have chosen to use neural network-based simulators to model object dynamics. Specifically, graph-based networks effectively learn the dynamics of various types of objects such as plasticine~\cite{shi2024robocraft, shi2023robocook}, cloth~\cite{pfaff2020learning, lin2022learning}, fluid~\cite{li2018learning, sanchez2020learning}, and stuffed animals~\cite{zhang2024dynamic}. GS-Dynamics~\cite{zhang2024dynamic} attempted to learn object dynamics directly from real-world videos using tracking and appearance priors from Dynamic Gaussians~\cite{luiten2024dynamic}, and generalized well to unseen actions. However, these learned models need extensive training samples and are often limited to specific environments with limited motion ranges. In contrast, our method requires only one interaction trial while achieving a broader range of motions.

\section{Preliminary: Spring-Mass Model}
\label{sec:preliminary}
Spring-mass models are widely used for simulating deformable objects due to their simplicity and computational efficiency. A deformable object is represented as a set of spring-connected mass nodes, forming a graph structure $\mathcal{G} = (\mathcal{V}, \mathcal{E})$, where $\mathcal{V}$ is the set of mass points and $\mathcal{E}$ is the set of springs. Each mass node $i$ has a position $\mathbf{x}_i \in \mathbb{R}^3$ and velocity $\mathbf{v}_i \in \mathbb{R}^3$, which evolve over time according to Newtonian dynamics. Springs are constructed between neighboring nodes based on a predefined topology, defining the elastic structure of the object.

The force on node $i$ is the result of the combined effects of adjacent nodes connected by springs:
\begin{equation}
\mathbf{F}_i = \sum_{(i, j) \in \mathcal{E}} \mathbf{F}_{i, j}^{\text{spring}} + \mathbf{F}_{i, j}^{\text{dashpot}} + \mathbf{F}_i^{\text{ext}},
\end{equation}
where the spring force and dashpot damping force between nodes $i$ and $j$ are given by $\mathbf{F}_{i, j}^{\text{spring}} = k_{ij} (\|\mathbf{x}_j - \mathbf{x}_i\| - l_{ij}) \frac{\mathbf{x}_j - \mathbf{x}_i}{\|\mathbf{x}_j - \mathbf{x}_i\|}$ and $\mathbf{F}_{i, j}^{\text{dashpot}} = -\gamma (\mathbf{v}_i - \mathbf{v}_j)$, respectively. Here, $k_{ij}$ is the spring stiffness, $l_{ij}$ is the rest length, and $\gamma$ is the dashpot damping coefficient. The external force $\mathbf{F}_i^{\text{ext}}$ accounts for factors such as gravity, collisions, and user interactions. The spring force restores the system to its rest shape, while the dashpot damping dissipates energy, preventing oscillations. For collisions, we use impulse-based collision handling when two mass points are very close, including collisions between the object and the collider, as well as between two object points.

The spring-mass model updates the system state with a dynamic model
$
\label{eq:transition}
\mathbf{X}_{t+1} = f_{\alpha, \mathcal{G}_0}(\mathbf{X}_{t}, a_t)
$
by applying explicit Euler integration to both velocity and position. More formally, for all $i$,
$\label{eq:state_update}
\mathbf{v}_i^{t+1} = \delta\left(\mathbf{v}_i^t + \Delta t\,\frac{\mathbf{F}_i}{m_i}\right), \quad
\mathbf{x}_i^{t+1} = \mathbf{x}_i^t + \Delta t\,\mathbf{v}_i^{t+1},$
where \(\mathbf{X}_t\) represents the system state at time \(t\), and $\delta$ represents the drag damping. In this formulation, \(\alpha\) denotes all physical parameters of the spring-mass model, including spring stiffness, collision parameters, and damping. It also encompasses the parameters related to the control interaction. \(\mathcal{G}_0\) represents the ``canonical'' geometry and topology for the spring-mass system\footnote{In practice, we use the first-frame object state as the canonical state.}, and \(a_t\) represents the action at time \(t\).

\section{Method}

In this section, we formulate the construction of \ourabbr as an optimization problem. We then present our two-stage strategy, where the first stage addresses the physics-related optimization, followed by the appearance-based optimization in the second stage. Finally, we demonstrate the capability of our framework to perform real-time simulation using the constructed \ourabbr.

\subsection{Problem Formulation}
\label{sec:ps}
Given three RGBD videos of a deformable object under interaction, our objective is to construct a \ourabbr model that captures the geometry, appearance, and physical parameters of the object over time. At each time frame \( t \), we denote the RGBD observations from the \( i \)-th camera as \( \mathbf{O}_{t,i} \), where \( \mathbf{O} = (\mathbf{I}, \mathbf{D}) \) represents the RGB image \( \mathbf{I} \) and depth map \( \mathbf{D} \).

The goal of our optimization problem is to minimize the discrepancy between the predicted observation \( \hat{\mathbf{O}}_{t,i} \) and the actual observation \( \mathbf{O}_{t,i} \). The predicted observation is derived by projecting and rendering the predicted state \( \hat{\mathbf{X}}_t \) onto images through a function \( g_\theta \), where $\theta$ encodes the appearance of the objects represented by Gaussian splats. The 3D state \( \hat{\mathbf{X}}_t \) evolves over time according to the Spring-Mass model, which captures the deformable object's dynamics and updates the state using the explicit Euler integration method. The optimization problem is formulated as:
\begin{equation}
\label{eq:goal}
\begin{aligned}
    \min_{\alpha, \mathcal{G}_0, \theta} \sum_{t, i} &C(\hat{\mathbf{O}}_{t, i}, \mathbf{O}_{t,i}) \\
    \text{s.t.} \quad \hat{\mathbf{O}}_{t, i} = g_\theta(\hat{\mathbf{X}}_{t}, i),  \quad&\quad \hat{\mathbf{X}}_{t+1} = f_{\alpha, \mathcal{G}} (\hat{\mathbf{X}}_{t}, a_t),
\end{aligned}
\end{equation}
where $\alpha, \mathcal{G}_0, \theta$ captures the physics, geometry, topology and appearance parameters (\cref{sec:preliminary}); the cost function quantifies the difference between the predicted observation \( \hat{\mathbf{O}}_{t,i} \) and the actual observation \( \mathbf{O}_{t,i} \). This cost function is decomposed into three components:
$
C = C_{\mathrm{geometry}} + C_{\mathrm{motion}} + C_{\mathrm{render}},
$
each capturing the discrepancy between the inferred system states and the corresponding observations from 3D geometry, 3D motion tracking, and 2D color, respectively (we defer the details of each cost component to Sec.~\ref{sec:invphysics} and Sec.~\ref{sec:gaussians}).
The function \( g_\theta \) is the observation model, describing the projection from the predicted state to the image plane and rendering image-space sensory observation from the $i$-th camera. The \( f_{\alpha, \mathcal{G}} \) models the dynamic evolution of the object’s state under the Spring-Mass model (\cref{sec:preliminary}).

\subsection{PhysTwin Framework}

Given the complexity of the overall optimization defined in Eq.~\ref{eq:goal}, our PhysTwin framework decomposes it into two stages. The first stage focuses on optimizing the geometry and physical parameters, while the second stage is dedicated to optimizing the appearance-related parameters.

\subsubsection{Physics and Geometry Optimization}
\label{sec:invphysics}
As outlined in our optimization formulation in \cref{sec:ps}, the objective is to minimize the discrepancy between the predicted observation \( \hat{\mathbf{O}}_{t,i} \) and the actual observation \( \mathbf{O}_{t,i} \). First, we convert the depth observations \( \mathbf{D}_t \) at each time frame \( t \) into the observed partial 3D point cloud \( \mathbf{X}_{t} \). 
In the first stage, we consider the following formulation for the optimization:
\begin{equation}
\label{eq:physics_and_geometry}
\begin{aligned}
    &\min_{\alpha, \mathcal{G}_0} \sum_t \left( C_{\text{geometry}}(\hat{\mathbf{X}}_t, \mathbf{X}_{t}) + C_{\text{motion}}(\hat{\mathbf{X}}_t, \mathbf{X}_{t}) \right) \\
    &\text{s.t.} \quad \hat{\mathbf{X}}_{t+1} = f_{\alpha, \mathcal{G}_0} (\hat{\mathbf{X}}_{t}, a_t),
\end{aligned}
\end{equation}
where the \( C_{\text{geometry}} \) function quantifies the single-direction Chamfer distance between the partially observed point cloud \( \mathbf{X}_{t} \) and the inferred state \( \hat{\mathbf{X}}_t \), and \( C_{\text{motion}} \) quantifies the tracking error between the predicted point \( \hat{\mathbf{x}}_i^t \) and its corresponding observed tracking \( \mathbf{x}_i^t \). The observed tracking is obtained using the vision foundation model CoTracker3~\cite{karaev2024cotracker3}, followed by lifting the result to 3D via depth map unprojection.

There are three main challenges in the first-stage optimization: (1) partial observations from sparse viewpoints; (2) joint optimization of both the discrete topology and physical parameters; and (3) discontinuities in the dynamic model, along with the long time horizon and dense parameter space, which make continuous optimization difficult. To address these challenges, we handle the geometry and other parameters separately. Specifically, we first leverage generative shape initialization to obtain the full geometry, then employ our two-stage sparse-to-dense optimization to refine the remaining parameters.

\textbf{Generative Shape Prior.} 
Due to partial observations, recovering the full geometry is challenging. We leverage a shape prior from the image-to-3D generative model TRELLIS~\cite{xiang2024structured} to generate a complete mesh conditioned on a single RGB observation of the masked object. To improve mesh quality, the input to TRELLIS is first enhanced using a super-resolution model~\cite{rombach2022high} that upscales the segmented foreground (obtained via Grounded-SAM2~\cite{ren2024grounded}). While the resulting mesh corresponds reasonably well with the camera observation, we can still observe inconsistencies in scale, pose, and deformation.

To address this, we design a registration module that uses 2D matching for scale estimation, rigid registration, and non-rigid deformation. A coarse-to-fine strategy first estimates initial rotation via 2D correspondences matched using SuperGlue~\cite{sarlin2020superglue}, followed by refinement with the Perspective-n-Point (PnP)~\cite{lepetit2009ep} algorithm. We resolve scale and translation ambiguities by optimizing the distances between matched points in the camera coordinate system. After applying these transformations, the objects are aligned in pose, with some deformations handled by as-rigid-as-possible registration~\cite{sorkine2007rigid}. Finally, ray-casting alignment ensures that observed points match the deformed mesh without occlusions.

These steps yield a shape prior aligned with the first-frame observations, which serves as a crucial initialization for the inverse physics and appearance optimization stages.

\textbf{Sparse-to-Dense Optimization.}
The Spring-Mass model consists of both the topological structure (i.e., the connectivity of the springs) and the physical parameters defined on the springs. As mentioned in \cref{sec:preliminary}, we also include control parameters to connect springs between control points and object points, defined by a radius and a maximum number of neighbors.
Similarly, for topology optimization, we employ a heuristic approach that connects nearest-neighbor points, also parameterized by a connection radius and a maximum number of neighbors, thereby controlling the density of the springs. 
To extract control points from video data, we utilize Grounded-SAM2~\cite{ren2024grounded} to segment the hand mask and CoTracker3~\cite{karaev2024cotracker3} to track hand movements. After lifting the points to 3D, we apply farthest-point sampling to obtain the final set of control points.  

All the aforementioned components constitute the parameter space we aim to optimize. The two main challenges are: (1) some parameters are non-differentiable (e.g., the radius and maximum number of neighbors); and (2) to represent a wide range of objects, we model dense spring stiffness, leading to a parameter space with tens of thousands of springs.

To address these challenges, we introduce a hierarchical sparse-to-dense optimization strategy. Initially, we employ zero-order, sampling-based optimization to estimate the parameters, which naturally circumvents the issue of differentiability. However, zero-order optimization becomes inefficient when the parameter space is too large. Therefore, in the first stage, we assume homogeneous stiffness, allowing the topology and other physical parameters to achieve a good initialization.
In the second stage, we further refine the parameters using first-order gradient descent, leveraging our custom-built differentiable spring-mass simulator. This stage simultaneously optimizes the dense spring stiffness and collision parameters.

Beyond the optimization strategy, we incorporate additional supervision by utilizing tracking priors from vision foundation models. We lift the 2D tracking prediction into 3D to obtain pseudo-ground-truth tracking data for the 3D points, which forms a crucial component of our cost function as mentioned in \cref{eq:physics_and_geometry}.

By integrating our optimization strategy with a cost function that leverages additional tracking priors, our PhysTwin framework can effectively and efficiently model the dynamics of diverse interactable objects from videos.

\subsubsection{Appearance Optimization}
\label{sec:gaussians}

For the second-stage appearance optimization, to model object appearance, we construct a set of static 3D Gaussian kernels parameterized by $\theta$, with each Gaussian defined by a 3D center position $\mu$, a rotation matrix represented by a quaternion $q \in \textbf{SO}(3)$, a scaling matrix represented by a 3D vector $s$, an opacity value $\alpha$, and color coefficients $c$. We optimize $\theta$ here via \begin{equation}
\label{eq:splats}
    \min_\theta \sum_{t, i} C_{\mathrm{render}}(\hat{\mathbf{I}}_{i, t}, \mathbf{I}_{i, t}) \text{\ s.t. } \hat{\mathbf{I}}_{i, t} = g_{\theta}(\hat{\mathbf{X}}_{t}, i),
\end{equation} where $\hat{\mathbf{X}}_{t}$ is the optimized system states at time $t$, $i$ is the camera index, and $\mathbf{I}_{i, t}$, $\hat{\mathbf{I}}_{i, t}$ are the ground truth image and rendered image from camera view $i$ at time $t$, respectively. $C_{\mathrm{render}}$ computes the $\mathcal{L}_1$ loss with a D-SSIM term between the rendering and ground truth image. For simplicity, we set $t=0$ to optimize appearance only at the first frame. We restrict the Gaussian shape to be isotropic to prevent spiky artifacts during deformation.

To ensure realistic rendering under deformation, we need to dynamically adjust each Gaussian at each timestep $t$ based on the transition between states $\hat{\mathbf{X}}_{t}$ and $\hat{\mathbf{X}}_{t+1}$. To achieve this, we adopt a Gaussian updating algorithm using Linear Blend Skinning (LBS)~\cite{sumner2007embedded, zhang2024dynamic, huang2024sc}, which interpolates the motions of 3D Gaussians using the motions of neighboring mass nodes. Please refer to the supplementary for details.

\subsection{Capabilities of \ourabbr}
Our constructed \ourabbr supports real-time simulation of deformable objects under various motions while maintaining realistic appearance. This real-time, photorealistic simulation enables interactive exploration of object dynamics. 

By introducing control points and dynamically connecting them to object points via springs, our system can simulate diverse motion patterns and interactions. These capabilities make \ourabbr a powerful representation for real-time interactive simulation and model-based robotic motion planning, which are further described in \cref{sec:application}.

\begin{figure*}
    \centering
    \vspace{-20pt}
    \includegraphics[width=0.98\linewidth]{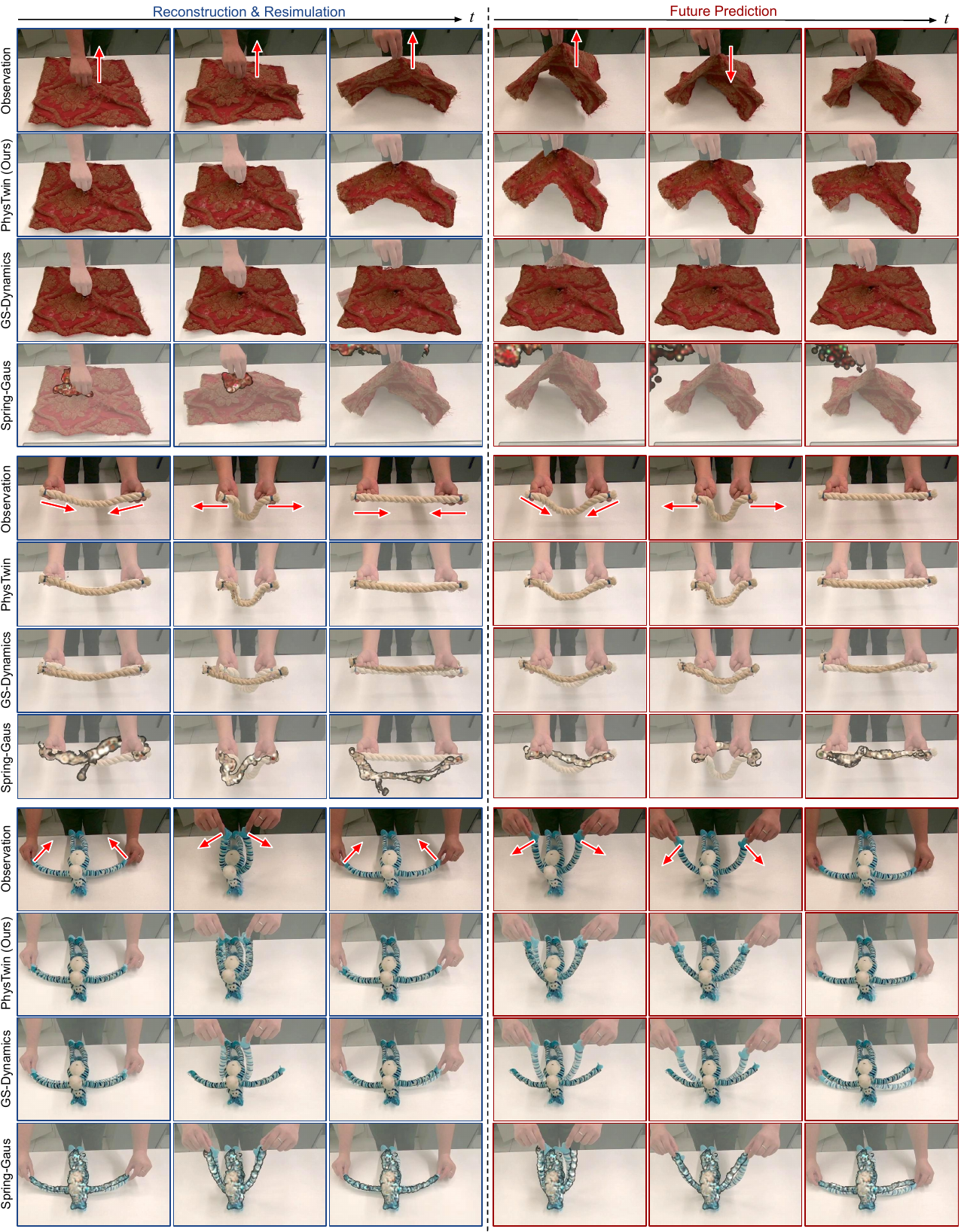}
    \vspace{-5pt}
    \captionof{figure}{\small
    \textbf{Qualitative Results on Reconstruction \& Resimulation and Future Prediction.} We visualize the rendering results of different methods on two tasks. For the reconstruction \& resimulation task, our method achieves a better match with the observations. For the future prediction task, our method accurately predicts the future state of the objects. In contrast, the baselines fail in most cases: GS-Dynamics~\cite{zhang2024dynamic} tends to remain static, while Spring-Gauss~\cite{zhong2024reconstruction} frequently causes the physical model to crash.
    }
    \label{fig:indomain}
\end{figure*}

\begin{figure*}
    \centering
    \vspace{-20pt}
    \includegraphics[width=\linewidth]{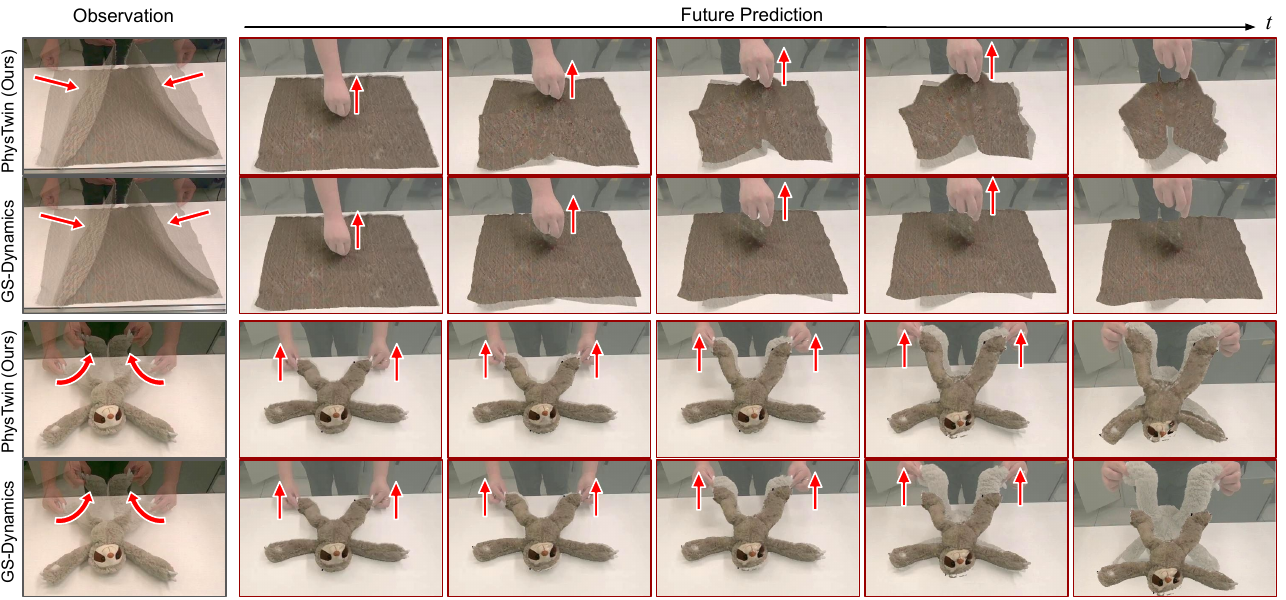}
    \vspace{-20pt}
    \captionof{figure}{\small
    \textbf{Qualitative Results on Generalization to Unseen Interactions.}
    We visualize the simulation of a deformable object under unseen interactions using our method and GS-Dynamics~\cite{zhang2024dynamic}. The leftmost image shows the interaction used to train the dynamics models, while the images on the right demonstrate their generalization to unseen interactions. Our PhysTwin significantly outperforms prior work.
    }
    \label{fig:outdomain}
\end{figure*}

\section{Experiments}

\begin{table*}[ht]
\caption{\small
    \textbf{Quantitative Results on Reconstruction \& Resimulation and Future Prediction.} We compare the performance of our method with two prior work, GS-Dynamics~\cite{zhang2024dynamic} and Spring-Gaus~\cite{zhong2024reconstruction}, on two tasks: reconstruction \& resimulation and future prediction. Our PhysTwin framework consistently outperforms the baselines across all metrics.}
\centering
\vspace{-5pt}
\resizebox{\linewidth}{!}{
\begin{tabular}{@{} l rr rr rr rr rr rr @{}}
\toprule
Task &  \multicolumn{6}{c}{Reconstruction \& Resimulation} & \multicolumn{6}{c}{Future Prediction} \\
\cmidrule(lr){2-7} \cmidrule(lr){8-13}
Method & CD $\downarrow$ & Track Error $\downarrow$ & IoU \% $\uparrow$ & PSNR $\uparrow$ & SSIM $\uparrow$ & LPIPS $\downarrow$ &  CD $\downarrow$ & Track Error $\downarrow$ & IoU \% $\uparrow$ & PSNR $\uparrow$ & SSIM $\uparrow$ & LPIPS $\downarrow$ \\
\midrule
Spring-Gaus~\cite{zhong2024reconstruction} & 0.041 & 0.050 & 57.6 & 23.445 & 0.928 & 0.102 & 0.062 & 0.094 & 46.4 & 22.488 & 0.924 & 0.113 \\
GS-Dynamics~\cite{zhang2024dynamic} & 0.014 & 0.022 & 72.1 & 26.260 & 0.940 & 0.052 & 0.041 & 0.070 & 49.8 & 22.540 & 0.924 & 0.097 \\
PhysTwin (\textbf{Ours}) & \textbf{0.005} & \textbf{0.009} & \textbf{84.4} & \textbf{28.214} & \textbf{0.945} & \textbf{0.034} & \textbf{0.012} & \textbf{0.022} & \textbf{72.5} & \textbf{25.617} & \textbf{0.941} & \textbf{0.055} \\
\bottomrule
\end{tabular}
}
\vspace{-10pt}
\label{tab:quant_indomain}
\end{table*}

\begin{table}[ht]
\caption{\small
    \textbf{Quantitative Results on Generalization to Unseen Interactions.} We compare our method with GS-Dynamics~\cite{zhang2024dynamic} on generalization to unseen interactions. Both methods are trained on the same video with a specific interaction and tested on unseen interactions. Our method achieves significantly better results.
}
\centering
\vspace{-5pt}
\resizebox{\linewidth}{!}{
\begin{tabular}{@{} l rr rr rr @{}}
\toprule
Method & CD $\downarrow$ & Track Error $\downarrow$ & IoU \% $\uparrow$ & PSNR $\uparrow$ & SSIM $\uparrow$ & LPIPS $\downarrow$ \\
\midrule
GS-Dynamics~\cite{zhang2024dynamic} & 0.029 & 0.038 & 63.4 & 25.053 & 0.934 & 0.067  \\
PhysTwin (\textbf{Ours}) & \textbf{0.013} & \textbf{0.018} & \textbf{72.18} & \textbf{26.199} & \textbf{0.938} & \textbf{0.047}\\
\bottomrule
\end{tabular}
}
\vspace{-10pt}
\label{tab:quant_outdomain}
\end{table}

In this section, we evaluate the performance of our PhysTwin framework across three distinct tasks involving different types of objects. Our primary objective is to address the following three questions:
(1)~How accurately does our framework reconstruct and resimulate deformable objects and predict their future states?  
(2)~How well does the constructed \ourabbr generalize to unseen interactions?
(3)~What is the utility of \ourabbr in downstream tasks?

\subsection{Experiment Settings}

\textbf{\indent Dataset.}
We collect a dataset of RGBD videos capturing human interactions with various deformable objects with different physical properties, such as ropes, stuffed animals, cloth, and delivery packages. Three RealSense-D455 RGBD cameras are used to record the interactions. 
Each video is 1 to 10 seconds long and captures different interactions, including quick lifting, stretching, pushing, and squeezing with one or both hands. 
We collect 22 scenarios encompassing various object types, interaction types, and hand configurations. For each scenario, the RGBD videos are split into a training set and a test set following a 7:3 ratio, where only the training set is used to construct \ourabbr. 
We manually annotate 9 ground-truth tracking points for each video to evaluate tracking performance with the semi-auto tool introduced in~\cite{doersch2023tapir}. 

\begin{figure*}
    \centering
    \vspace{-20pt}
    \includegraphics[width=\linewidth]{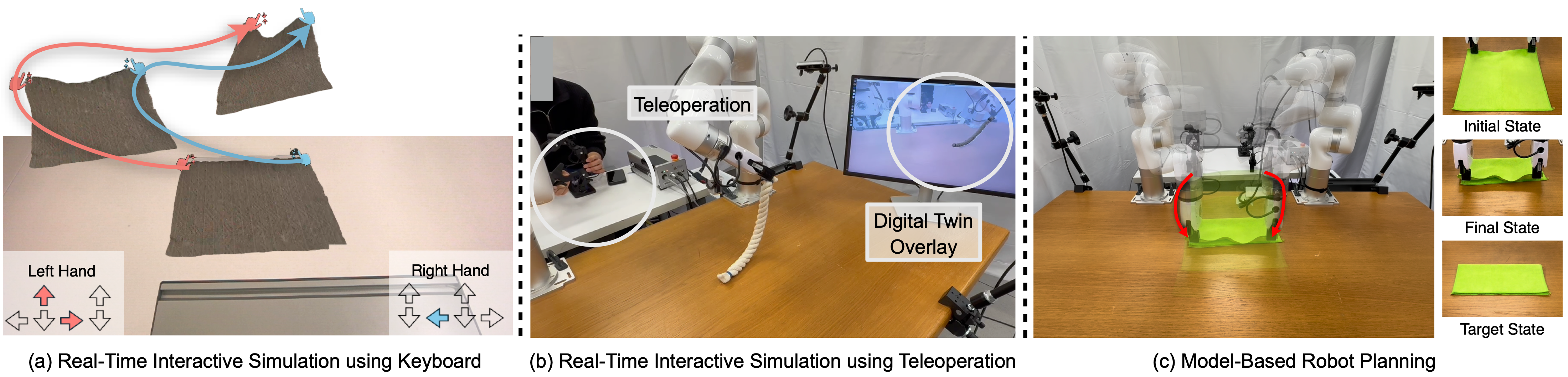}
    \vspace{-22pt}
    \captionof{figure}{\small
    \textbf{Applications of our \ourabbr.} Our constructed PhysTwin supports a variety of tasks, including real-time interactive simulation, which can accept input from either a keyboard or a robot teleoperation setup. Meanwhile, PhysTwin also enables model-based robot planning to accomplish tasks such as lifting a rope into some specific configuration. }
    \label{fig:application}
    \vspace{-13pt}
\end{figure*}

\textbf{Tasks.}
To assess the effectiveness of our PhysTwin framework and the quality of our constructed \ourabbr, we formulate three tasks:  
(1)~Reconstruction \& Resimulation;  
(2)~Future Prediction; and 
(3)~Generalization to Unseen Actions.

For the Reconstruction \& Resimulation task, the objective is to construct \ourabbr such that it can accurately reconstruct and resimulate the motion of deformable objects given the actions represented by the control point positions. 

For the Future Prediction task, we aim to assess whether \ourabbr can perform well on unseen future frames during its construction.  
For the Generalization to Unseen Interactions task, the goal is to assess whether \ourabbr can adapt to different interactions. To evaluate this, we construct a generalization dataset consisting of interaction pairs performed on the same object but with varying motions, including differences in hand configuration and interaction type.

\textbf{Baselines.}
To the best of our knowledge, there is currently no existing work that demonstrates good performance across all three tasks. Therefore, we select two main research directions as baselines and further augment them to match the tasks in our setting (full details in the supplementary).

The first baseline we consider is a physics-based simulation method for identifying the material properties of deformable objects, Spring-Gaus~\cite{zhong2024reconstruction}. Their work has demonstrated strong capabilities in reconstruction, resimulation, and future prediction in its original setting. However, their framework does not support external control inputs, so we augment it with additional control capabilities.

The second baseline is a learning-based simulation approach, GS-Dynamics~\cite{zhang2024dynamic}, which employs a GNN-based neural dynamics model to learn system dynamics directly from partial observations. In their original setting, video preprocessing with Dyn3DGS~\cite{luiten2024dynamic} is required to obtain tracking information. For a fairer comparison, we strengthened it by using our 3D-lifting tracker based on CoTracker3~\cite{karaev2024cotracker3}, which provides more efficient and accurate supervision for training the neural dynamics model used by GS-Dynamics.

\textbf{Evaluation.}
To better understand whether our prediction matches the observations, we evaluate predictions in both 3D and 2D. For the 3D evaluation, we use the single-direction Chamfer Distance (partial ground truth with our full-state prediction) and the tracking error (based on our manually annotated ground-truth tracking points). For the 2D evaluation, we assess image quality using PSNR, SSIM, and LPIPS~\cite{zhang2018perceptual}, and silhouette alignment using IoU. We perform 2D evaluation only at the center viewpoint due to optimal visibility of objects, with metrics averaged across all frames and scenarios. Specially, for the Spring-Gaus~\cite{zhong2024reconstruction} baseline, its optimization process is unstable due to inaccurate physics modeling. Therefore, we report the above metrics only for its successful cases.

\subsection{Results}

To assess the performance of our framework and the quality of our constructed \ourabbr,
we compare with two augmented baselines across three task settings. Our quantitative analysis reveals that the PhysTwin framework consistently outperforms the baselines across various tasks. 

\textbf{Reconstruction \& Resimulation.}  
The quantitative results in~\cref{tab:quant_indomain} Reconstruction \& Resimulation column demonstrate the superior performance of our PhysTwin method over baselines. Our approach significantly improves all evaluated metrics, including Chamfer Distance, tracking error, and 2D IoU, confirming that our reconstruction and resimulation align more closely with the original observations. This highlights the effectiveness of our model in learning a more accurate dynamics model under sparse observations.
Additionally, rendering metrics show that our method produces more realistic 2D images, benefiting from the Gaussian blending strategy and enhanced dynamic modeling. \Cref{fig:indomain} further provides qualitative visualizations across different objects, illustrating precise alignment with original observations.
Notably, our physics-based representation inherently improves point tracking. After physics-constrained optimization, our tracking surpasses the original CoTracker3~\cite{karaev2024cotracker3} predictions used for training, achieving better alignment after global optimization (See supplement for more details).

\textbf{Future Prediction.}  
\Cref{tab:quant_indomain}, in the Future Prediction column, demonstrates that our method achieves superior performance in predicting unseen frames, excelling in both dynamics alignment and rendering quality. \Cref{fig:indomain} further provides qualitative results, illustrating the accuracy of our predictions on unseen frames.

\textbf{Generalization to Unseen Interactions.}  
We also evaluate the generalization performance to unseen interactions. 
Our dataset includes transfers from one interaction (e.g., single lift) to significantly different interactions (e.g., double stretch). 
We directly use our constructed \ourabbr and leverage our registration pipeline to align it with the first frame of the target case. \Cref{fig:outdomain} shows that our method closely matches the ground truth observations in terms of dynamics. Quantitative results further demonstrate the robustness of our method across different actions. In contrast, the neural dynamics model struggles to adapt to environmental changes and diverse interactions as effectively as our approach. Moreover, in unseen interaction scenarios, our method achieves performance comparable to that on the future prediction task, highlighting the robustness and generalization capability of our constructed PhysTwin.

\subsection{Application}
\label{sec:application}
The efficient forward simulation capabilities of our Spring-Mass simulator, implemented using Warp~\cite{warp2022}, enable a variety of downstream applications.
\Cref{fig:application} showcases key applications enabled by our \ourabbr:
(1)~Interactive Simulation: Users can interact with objects in real time using keyboard controls, either with one or both hands. The system also supports real-time simulation of an object's future state during human teleoperation with robotic arms. This feature serves as a valuable tool for predicting object dynamics during manipulation.
(2)~Model-Based Robotic Planning: Owing to the high fidelity of our constructed \ourabbr, it can be used as a dynamic model in planning pipelines. By integrating it with model-based planning techniques, we can generate effective motion plans for robots to complete a variety of tasks.

\section{Conclusion}
We introduced PhysTwin, a novel framework for constructing physical digital twins from sparse videos, enabling effective reconstruction and resimulation of deformable objects. Our approach excels in predicting future states and simulating object interactions that generalize to unseen actions. 
We showed the superior performance of our method across various object types, control configurations, and task settings, significantly outperforming prior work. 
\ourabbr enables various downstream tasks that demand high-speed simulation and accurate future prediction. 
Moreover, our approach provides valuable insights for robotic manipulation. By bridging perception and physics-based simulation, \ourabbr serves as a crucial tool for guiding robot interactions, making real-world deployment more efficient and reliable.

\section*{Acknowledgement}
This work is partially supported by the Toyota Research Institute (TRI), the Sony Group Corporation, Google, Dalus AI, the DARPA TIAMAT program (HR0011-24-9-0430), the Intel AI SRS gift, Amazon-Illinois AICE grant, Meta Research Grant, IBM IIDAI Grant, and NSF Awards \#2331878, \#2340254, \#2312102, \#2414227, and \#2404385. We greatly appreciate the NCSA for providing computing resources. This article solely reflects the opinions and conclusions of its authors and should not be interpreted as necessarily representing the official policies, either expressed or implied, of the sponsors.

{
    \small
    \bibliographystyle{ieeenat_fullname}
    \bibliography{main}
}

\newpage
\appendix
\addtocontents{toc}{\setcounter{tocdepth}{3}}
\setlength{\cftbeforesecskip}{2pt}
\renewcommand{\contentsname}{Supplement Index}
{
  \hypersetup{linkcolor=black}
  \tableofcontents
}

\vspace{5pt}

In the supplement, we provide additional details of our PhysTwin framework, more qualitative results across different tasks, and further analysis of our methods. All the videos showcasing our results on various instances, interactions, and tasks are available on our website.

\section{Additional Details for the Shape Prior}
As mentioned in the main paper, we leverage TRELLIS \cite{xiang2024structured} to generate the full mesh from a single RGB observation. However, the potential non-rigid registration presents a non-trivial challenge.

To address these issues, we design a registration module that leverages 2D matching to handle scale estimation, rigid registration, and non-rigid deformation. First, to estimate the initial rotation, we adopt a coarse-to-fine strategy. We use uniformly distributed virtual cameras placed on a sphere surrounding the object to render images and match 2D correspondences using SuperGlue \cite{sarlin2020superglue}. Based on the number of matches, we select the view with the maximum number of correspondences, providing a rough rotation estimate. We then apply the Perspective-n-Point (PNP) algorithm to refine the 3D matched points on the generated mesh and the corresponding 2D pixels in the observation, estimating the precise rotation matrix.

After estimating the rotation, translation and scale ambiguities may still exist. To resolve these, we optimize the distances between matched point pairs to solve for scale and translation. This is simplified in the camera coordinate system, as after PNP, the matched points in the generated mesh and the corresponding points in the real observation point cloud lie along the same line connecting the origin. Therefore, the scale and translation optimization can be reduced to optimizing only the scale. Once these transformations are applied, the two objects should be in similar poses, with some parts undergoing non-rigid deformations. To handle such deformations, we use an as-rigid-as-possible registration to deform the mesh into a non-rigid pose matching the real observation. Finally, we perform ray-casting alignment, shooting rays from the camera to ensure that the observed points align with the deformed mesh and are neither occluded nor occlude the mesh.

\begin{table*}[ht]
\caption{\small
    \textbf{Ablations of Our Sparse-to-Dense Optimization.} To better understand our optimization process, we conduct ablation experiments comparing results with only zero-order optimization or first-order optimization. The results demonstrate that our sparse-to-dense optimization strategy is effective in obtaining the most accurate physical parameters.
}
\centering
\resizebox{\linewidth}{!}{
\begin{tabular}{@{} l rr rr rr rr rr rr @{}}
\toprule
Task &  \multicolumn{6}{c}{Reconstruction \& Resimulation} & \multicolumn{6}{c}{Future Prediction} \\
\cmidrule(lr){2-7} \cmidrule(lr){8-13}
Method & CD $\downarrow$ & Track Error $\downarrow$ & IoU \% $\uparrow$ & PSNR $\uparrow$ & SSIM $\uparrow$ & LPIPS $\downarrow$ &  CD $\downarrow$ & Track Error $\downarrow$ & IoU \% $\uparrow$ & PSNR $\uparrow$ & SSIM $\uparrow$ & LPIPS $\downarrow$ \\
\midrule
Zero-order Only & 0.007 & 0.012 & 80.2 & 27.409 & 0.943 & 0.039 & 0.014 & 0.025 & 69.2 & 25.008 & 0.938 & 0.061 \\
First-order Only & 0.008 & 0.012 & 82.7 & 27.913 & 0.944 & 0.037 & 0.019 & 0.034 & 65.7 & 24.572 & 0.936 & 0.067 \\
PhysTwin (\textbf{Ours}) & \textbf{0.005} & \textbf{0.009} & \textbf{84.4} & \textbf{28.214} & \textbf{0.945} & \textbf{0.034} & \textbf{0.012} & \textbf{0.022} & \textbf{72.5} & \textbf{25.617} & \textbf{0.941} & \textbf{0.055} \\
\bottomrule
\end{tabular}
}
\label{tab:quant_ablation}
\end{table*}

\begin{figure}
    \centering
    \includegraphics[width=\linewidth]{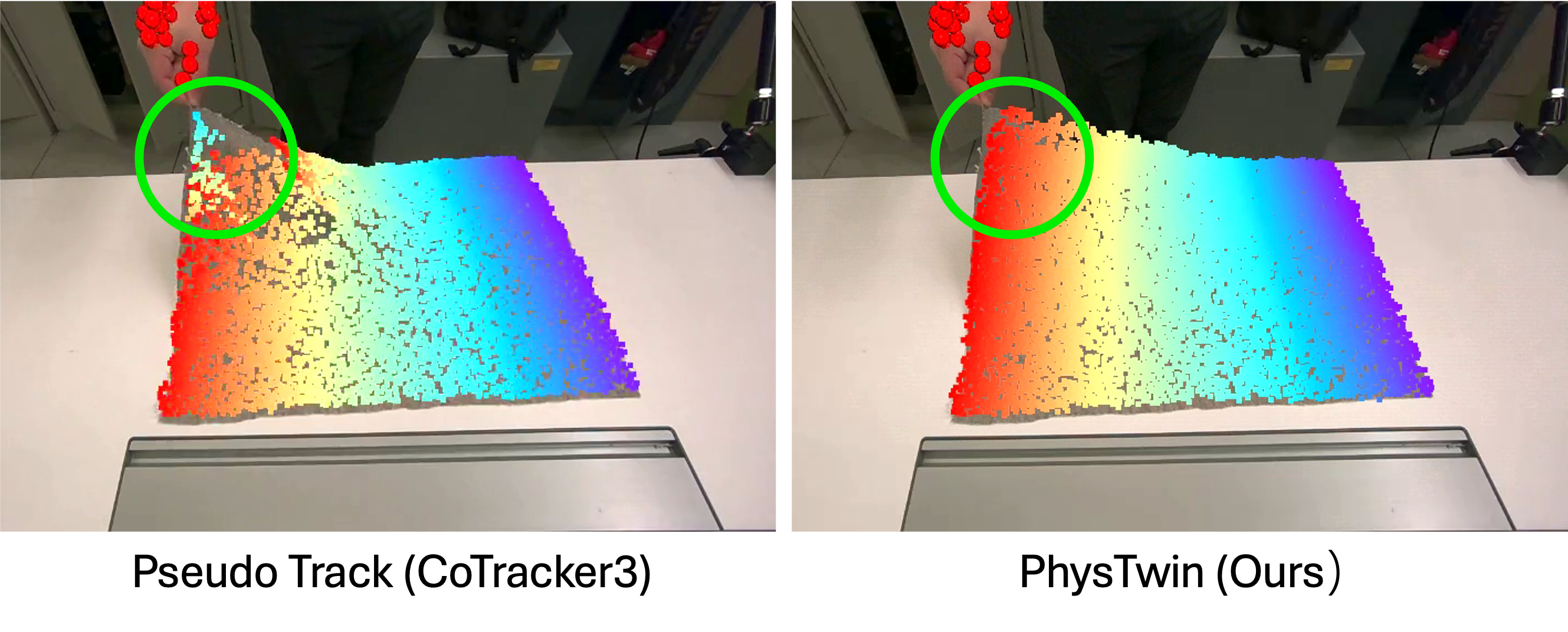} 
    \vspace{-15pt}
    \captionof{figure}{\small
    \textbf{Visualization of Tracking Results.} We compare the tracking results produced by our PhysTwin with the raw tracking results from CoTracker3 \cite{karaev2024cotracker3}. Our PhysTwin achieves more natural and smoother movement compared to the raw predictions from CoTracker3.
    }
    \label{fig:tracking}
\end{figure}

\begin{figure*}
    \centering
    \vspace{-20pt}
    \includegraphics[width=0.98\linewidth]{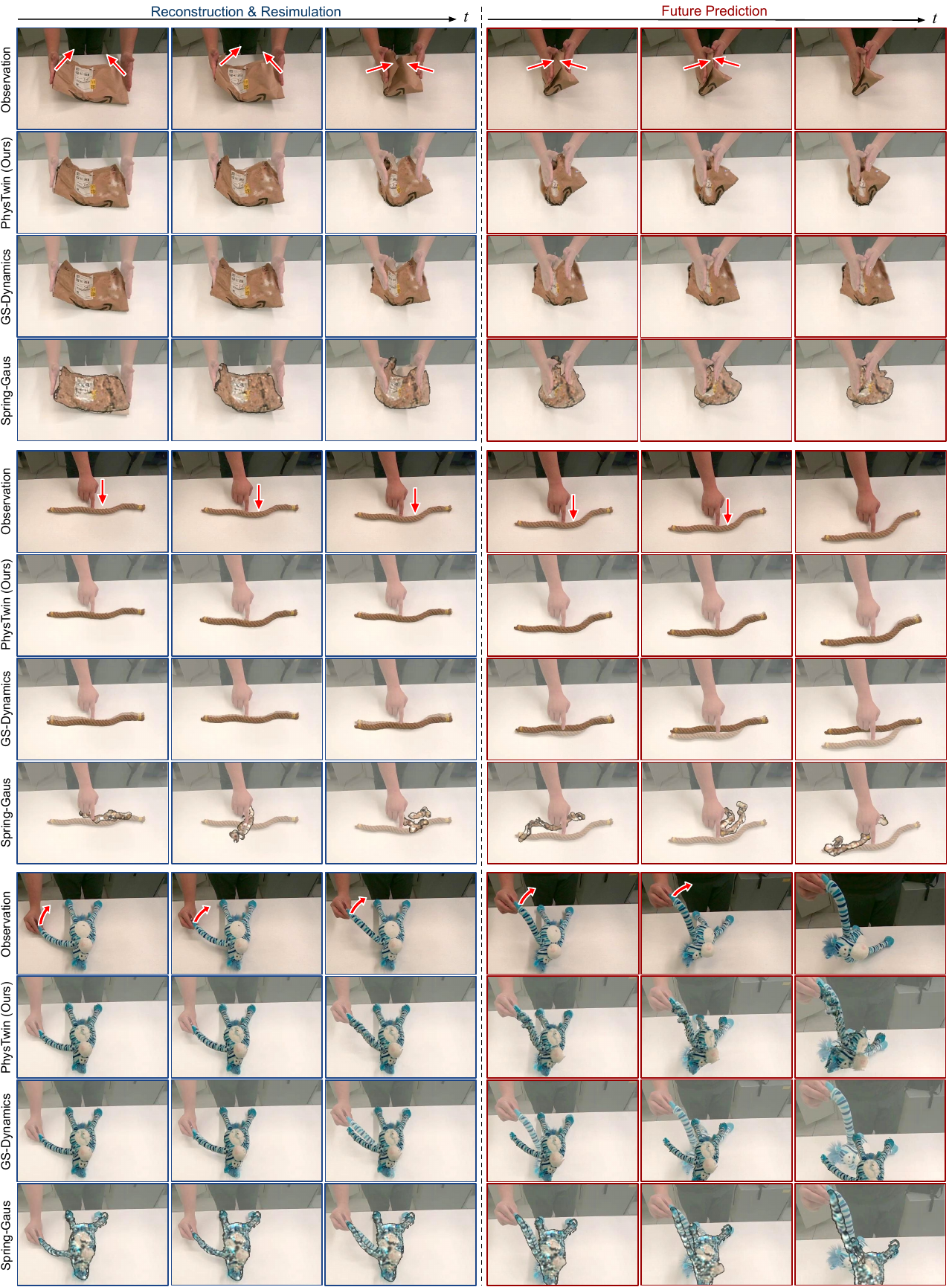}
    \vspace{-5pt}
    \captionof{figure}{\small
    \textbf{Additional Qualitative Results on Reconstruction \& Resimulation and Future Prediction.}
    }
    \label{fig:indomain_supp}
\end{figure*}

\begin{figure*}
    \centering
    \includegraphics[width=\linewidth]{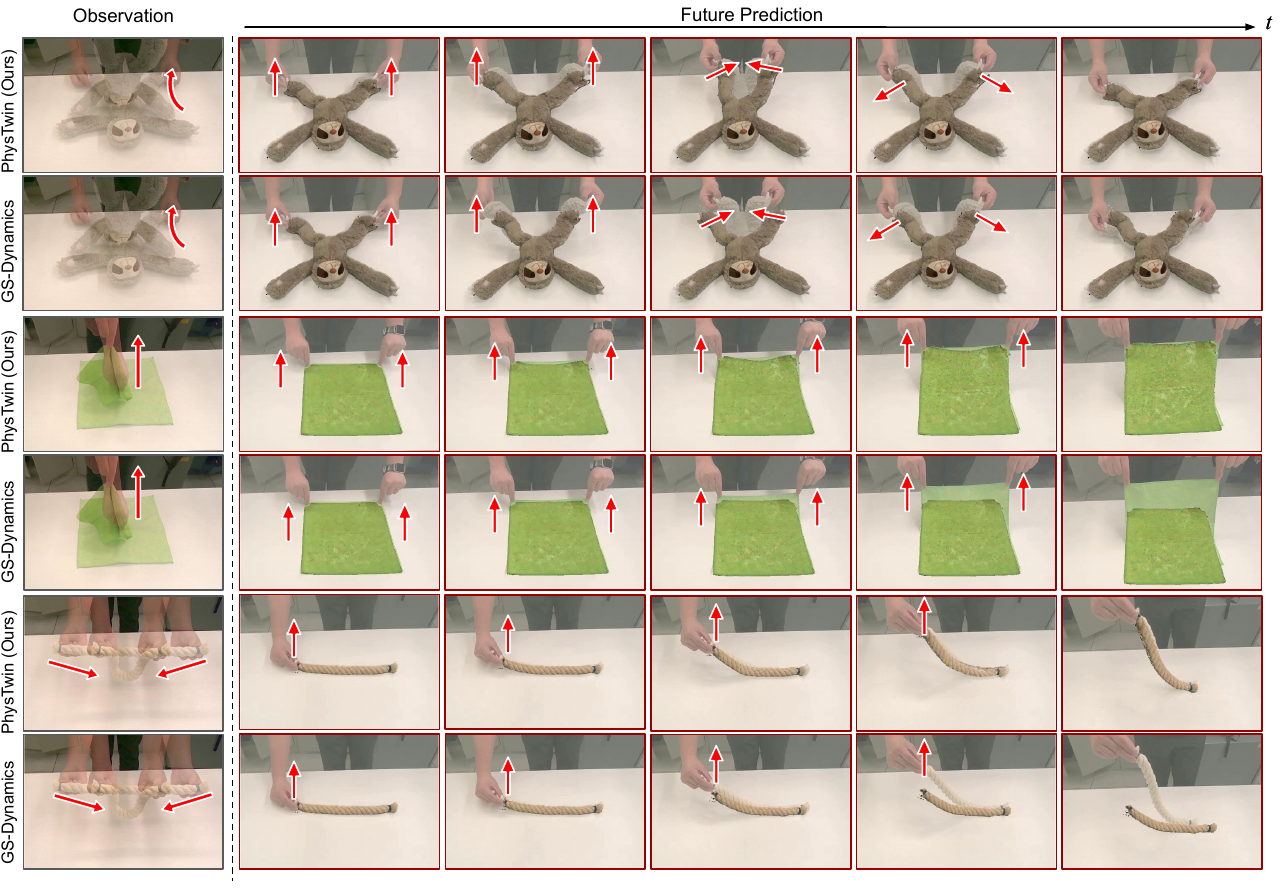}
    \vspace{-25pt}
    \captionof{figure}{\small
    \textbf{Additional Qualitative Results on Generalization to Unseen Interactions.} 
    }
    \label{fig:outdomain_supp}
\end{figure*}

\begin{figure*}
    \centering
    \vspace{-20pt}
    \includegraphics[width=0.98\linewidth]{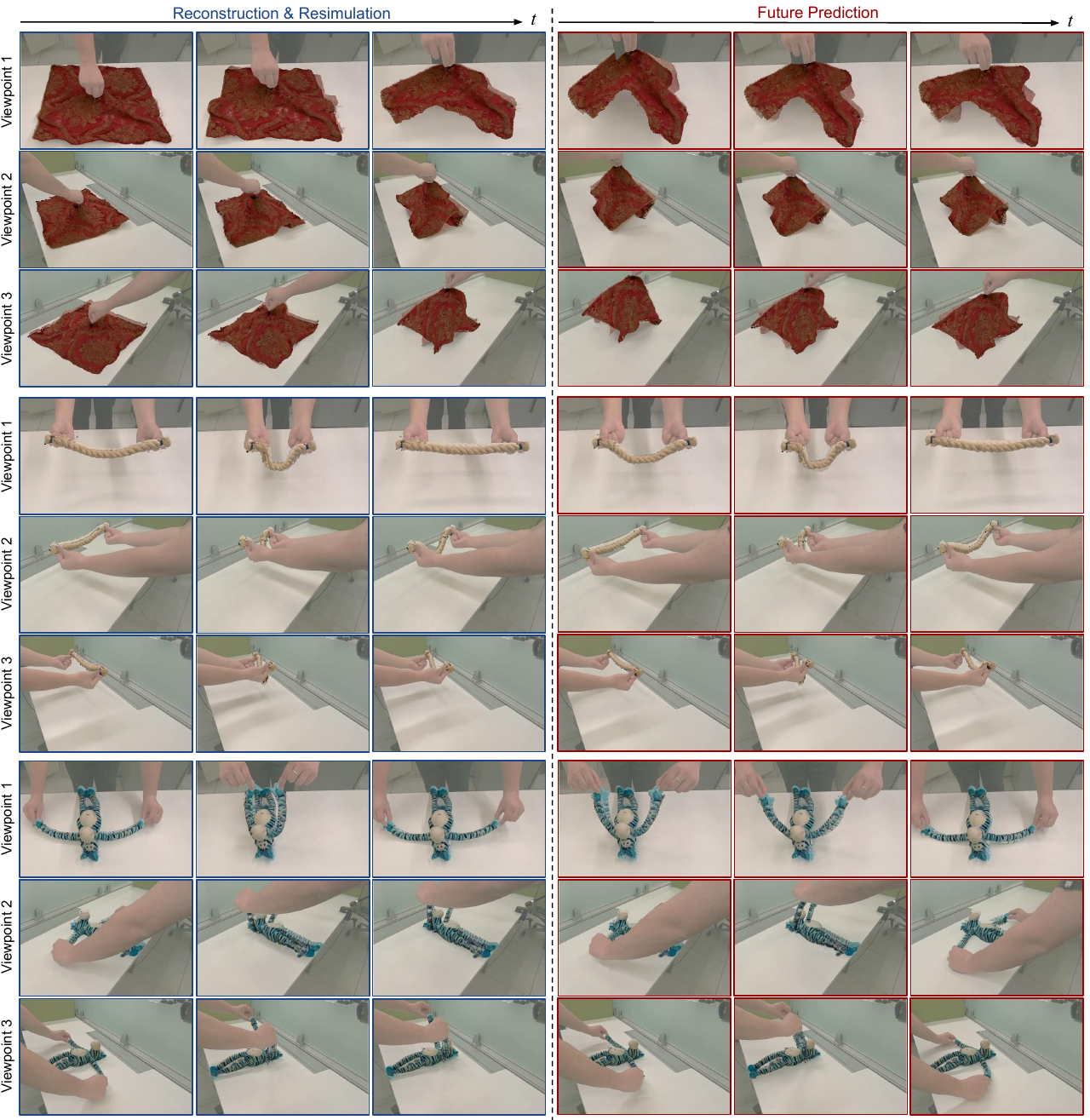}
    \vspace{-5pt}
    \captionof{figure}{\small
    \textbf{Qualitative Results on Reconstruction \& Resimulation and Future Prediction with different viewpoints.} 
    }
    \label{fig:vp_supp}
\end{figure*}

\section{Additional Details for 3D Gaussian Update}
Given the previous state $\hat{\mathbf{X}}_{t}$ and the predicted state $\hat{\mathbf{X}}_{t+1}$, we first solve for the 6-DoF transformation of each mass node $\hat{\mu}^t_i \in \hat{\mathbf{X}}_{t}$. For 3D translations, we obtain them from the predicted node translations $T^t_i$. For 3D rotations, for each vertex $\hat{\mu}^t_i$, we estimate a rigid local rotation $R^t_i$ based on motions of its neighbors $\mathcal{N}(i)$ from time $t$ to $t+1$:
\begin{equation}
R_i^t = \arg \min_{R \in SO(3)} \sum_{j \in \mathcal{N}(i)} \|R(\hat{\mu}_j^t - \hat{\mu}_i^t) - (\hat{\mu}_j^{t+1} - \hat{\mu}_i^{t+1})\|^2.
\end{equation}
In the next step, we transform Gaussian kernels using Linear Blend Skinning (LBS)~\cite{sumner2007embedded, zhang2024dynamic, huang2024sc} by locally interpolating the transformations of their neighboring nodes. Specifically, for the 3D center and rotation of each Gaussian:
\begin{equation}
\mu_j^{t+1} = \sum_{k \in \mathcal{N}(j)} w_{jk}^t (R_k^t (\mu_j^t - \hat{\mu}_k^t) + \hat{\mu}_k^t + T_k^t)
\end{equation}
\begin{equation}
q_j^{t+1} = (\sum_{k \in \mathcal{N}(j)} w_{jk}^t r_k^t) \otimes q_j^t,
\end{equation}
where $R_k^t \in \mathbb{R}^{3\times 3}$ and $r_k^t \in \mathbb{R}^{4}$ are the matrix and quaternion forms of the rotation of vertex $k$; $\otimes$ denotes the quaternion multiply operator; $\mathcal{N}(j)$ represents $K$-nearest vertices of a Gaussian center $\mu_j^t$; $w_{jk}^t$ is the interpolation weights between a Gaussian $\mu_j^t$ and a corresponding vertex $\hat{\mu}_k^t$, which are derived inversely proportional to their 3D distance:
\begin{equation}
w_{jk}^t = \frac{\|\mu_j^t - \hat{\mu}_k\|^{-1}}{\sum_{k \in \mathcal{N}(j)} \|\mu_j^t - \hat{\mu}_k\|^{-1}}
\end{equation}
to ensure larger weights are assigned to the spatially closer pairs. Finally, with the updated Gaussian parameters, we are able to perform rendering at timestep $t+1$ with the transformed 3D Gaussians.

\section{Additional Experimental Details}
Due to the page limit in the main paper, we provide additional qualitative results on different instances under various interactions, as well as further analysis experiments.

\textbf{Baselines.} As described in the main paper, we select two prior works for comparison: Spring-Gaus~\cite{zhong2024reconstruction} and GS-Dynamics~\cite{zhang2024dynamic}.  

For Spring-Gaus, while it demonstrates reasonable performance in modeling object-collision videos, its applicability is limited to relatively simple cases where objects primarily deform under gravity, restricting the range of supported object types. To adapt Spring-Gaus~\cite{zhong2024reconstruction} to our setting, we extend it by introducing support for control points. Specifically, we add additional springs that connect the control points to their neighboring object points within a predefined distance, enabling direct optimization on our dataset.  
Furthermore, to ensure compatibility with our sparse-view setup, we incorporate our shape prior as the initialization for their static Gaussian construction. Since their constructed Gaussians lack the ability to generalize to different initial conditions, we evaluate their approach only on the first two tasks: reconstruction \& resimulation and future prediction.

For GS-Dynamics, we compare our method with theirs across all three tasks.  
To enable the GNN-based dynamics model to produce realistic renderings, we augment it with our Gaussian blending strategy, enhancing its ability to generate high-quality images.  

\textbf{Tasks.} \ourabbr is constructed solely from the training set of each data point, and its performance is evaluated based on how well it matches the original video within the test set. For the generalization task, we create a dataset consisting of interaction pairs performed on the same object. For example, we construct \ourabbr for a sloth toy based on a scenario where it is lifted with one hand and then evaluate its performance in a different scenario where its legs are stretched using both hands. The dataset includes 11 such pairs, and since each pair allows for two possible transfer directions (i.e., from one interaction to another or vice versa), this results in a total of 22 generalization experiments. In this task, \ourabbr is still constructed using only the training set of the source interaction but is applied across the entire sequence of the target interaction.

\textbf{Qualitative Results.} We present more qualitative results for different instances across various interactions on our three tasks: reconstruction \& resimulation, future prediction (\cref{fig:indomain_supp}), and generalization to unseen interactions (\cref{fig:outdomain_supp}). All results demonstrate the superior performance of our method compared to prior work.

\textbf{Different Viewpoints.} \Cref{fig:vp_supp} presents the visualization of the rendering results from different viewpoints, demonstrating the robustness of our \ourabbr in handling various viewpoints.

\textbf{Ablation Study on Hierarchical Optimization.}
To better understand the importance of our hierarchical sparse-to-dense optimization strategy, we conduct ablation studies with two variants: one using only zero-order optimization and the other using only first-order optimization. These experiments are performed on both the reconstruction \& resimulation task and the future prediction task. \Cref{tab:quant_ablation} presents the results of different variants.
Our complete pipeline achieves the best performance across both tasks. The variant with only zero-order optimization fails to capture fine-grained material properties, limiting its ability to represent different objects. On the other hand, the variant with only first-order dense optimization neglects the optimization of non-differentiable parameters, such as the spring connections. The default connections fail to accurately model the real object structure, and the connection distances between control points and object points cannot be effectively handled with a fixed initialization value.

\textbf{Tracking Results.} \Cref{fig:tracking} shows the visualization of our tracking results and the pseudo-GT tracking results from CoTracker3 \cite{karaev2024cotracker3}. Even though our PhysTwin is optimized with noisy GT tracking, our model achieves much better and smoother tracking results during both the reconstruction \& resimulation and future prediction tasks.

\textbf{Data Efficiency Experiment.}  
To further analyze the performance difference between our method and the GNN-based approach, we collected 29 additional data points on the same motion (double-hand stretching and folding rope), bringing the total to 30 data points for training the neural dynamics model. In contrast, our method is trained using only 1 data point. The results show that GS-Dynamics does not show a performance boost even with 30 times more data than our method. This indicates that their approach is data-hungry, whereas our method demonstrates significantly better data efficiency in learning a useful dynamics model. Even with 30 times more data, the learning-based method still struggles to capture precise dynamics as effectively as our approach.

\section{Future Work}
Our work takes an important step towards constructing an effective physical digital twin for deformable objects from sparse video observations. Unlike existing methods that primarily focus on geometric reconstruction, our approach integrates physical properties, enabling accurate resimulation, future prediction, and generalization to unseen interactions. 
Despite using three RGBD views in our current setup, our framework is inherently flexible and can extend to even sparser observations. With appropriate priors, a single RGB video could serve as a promising and scalable alternative, making our approach more applicable to in-the-wild scenarios.
Furthermore, while our framework optimizes physical parameters based on a single type of interaction, expanding to multiple action modalities could further enhance the estimation of an object's intrinsic properties. Learning from a broader range of interactions may reveal richer physical characteristics and improve robustness.
Beyond reconstruction and resimulation, our method opens up exciting possibilities for downstream applications, particularly in robotics. By providing a structured yet efficient digital twin, our approach significantly simplifies real-to-sim transfer, reducing the reliance on domain randomization for reinforcement learning. Additionally, the high-speed simulation and real-time rendering capabilities of our framework pave the way for more effective model-based robotic planning.
By bridging the gap between perception and physics-based simulation, our method lays a solid foundation for future advancements in both computer vision and robotics.

\end{document}